\documentclass{article}

\usepackage{microtype}
\usepackage{graphicx}
\usepackage{subfigure}
\usepackage{booktabs} 

\usepackage{hyperref}

\usepackage[accepted]{icml2025}

\usepackage{amsmath}
\usepackage{amssymb}
\usepackage{mathtools}
\usepackage{amsthm}

\usepackage[capitalize,noabbrev]{cleveref}

\usepackage{multirow}
\usepackage{pifont}  
\usepackage{utfsym}
\usepackage{afterpage}  
\usepackage{tikz}
\usepackage[algo2e]{algorithm2e}
\usepackage{wrapfig}
\usepackage{tabularx}
\usepackage{tcolorbox}

\SetKwComment{Comment}{\color{green!50!black}\# }{}

\SetKwProg{Function}{def}{:}{}

\SetKwProg{For}{for}{:}{}
\SetKwProg{If}{if}{:}{}

\theoremstyle{plain}

\theoremstyle{definition}

\theoremstyle{remark}

\definecolor{mygray}{gray}{.9}
\definecolor{ggray}{RGB}{127,127,127}
\definecolor{reda}{RGB}{192,0,0}
\definecolor{redb}{RGB}{217,148,143}
\definecolor{myyellow}{RGB}{190,144,0}
\definecolor{mygreen}{RGB}{80,100,40}
\definecolor{myblue}{RGB}{30,90,100}

\usepackage{xspace}
\makeatletter
\DeclareRobustCommand\onedot{\futurelet\@let@token\@onedot}
\def\@onedot{\ifx\@let@token.\else.\null\fi\xspace}
\def\eg{\emph{e.g}\onedot} 
\def\ie{\emph{i.e}\onedot} 
\def\cf{\emph{c.f}\onedot}

\def\etal{\emph{et al}\onedot}
\newcommand{\thickhline}{
    \noalign {\ifnum 0=`}\fi \hrule height 1pt
    \futurelet \reserved@a \@xhline
}

\usepackage{algorithm}
\usepackage{algorithmic}
\usepackage{listings} 
\usepackage{colortbl}  
\usepackage{makecell}
\usepackage{enumitem}
\usepackage{tikz}
\usepackage{tcolorbox}
\usepackage{varwidth}
\usetikzlibrary{decorations.pathreplacing}
\definecolor{codegreen}{RGB}{79,126,127}
\definecolor{codedefine}{RGB}{153,54,159}
\definecolor{codefunc}{RGB}{73,122,234}
\definecolor{codecall}{RGB}{73,122,234}
\definecolor{codepro}{RGB}{212,96,80}
\definecolor{codedim}{RGB}{89,152,195}
\definecolor{mybrown}{RGB}{165,42,42}
\newcommand{\myhyperlink}[3][black]{\hyperlink{#2}{\color{#1}{#3}}}
\definecolor{dkgreen}{rgb}{0,0.6,0}
\definecolor{gray}{rgb}{0.5,0.5,0.5}
\definecolor{mauve}{rgb}{0.58,0,0.82}
\usepackage{algorithm}
\usepackage{algorithmic}
\usepackage{listings} 
\usepackage{colortbl}  
\usepackage{makecell}
\definecolor{codedefine}{RGB}{153,54,159}
\definecolor{codefunc}{RGB}{73,122,234}
\definecolor{codecall}{RGB}{73,122,234}
\definecolor{codepro}{RGB}{212,96,80}
\definecolor{codedim}{RGB}{89,152,195}

\definecolor{dkgreen}{rgb}{0,0.6,0}
\definecolor{gray}{rgb}{0.5,0.5,0.5}
\definecolor{mauve}{rgb}{0.58,0,0.82}

\usepackage[textsize=tiny]{todonotes}

\icmltitlerunning{Taking A Closer Look at Interacting Objects: Interaction-Aware Open Vocabulary Scene Graph Generation}

\begin{document}

\twocolumn[
\icmltitle{Taking A Closer Look at Interacting Objects: \\ Interaction-Aware Open Vocabulary Scene Graph Generation}



\icmlsetsymbol{equal}{*}

\begin{icmlauthorlist}
\icmlauthor{Lin Li}{hkust}
\icmlauthor{Chuhan Zhang}{hkust}
\icmlauthor{Dong Zhang}{hkust}
\icmlauthor{Chong Sun}{tencent}
\icmlauthor{Chen Li}{tencent}
\icmlauthor{Long Chen}{hkust}
\end{icmlauthorlist}

\icmlaffiliation{hkust}{The Hong Kong University of Science and Technology, Hong Kong}
\icmlaffiliation{tencent}{Tencent, China}

\icmlcorrespondingauthor{Long Chen}{longchen@ust.hk}


\vskip 0.3in
]



\printAffiliationsAndNotice{}  

\begin{abstract}
Today's open vocabulary scene graph generation (OVSGG) extends traditional SGG by recognizing novel objects and relationships beyond predefined categories, leveraging the knowledge from pre-trained large-scale models. Most existing methods adopt a two-stage pipeline: $\!$weakly supervised pre-training with image captions and supervised fine-tuning (SFT) on fully annotated scene graphs. Nonetheless, they omit explicit modeling of \textit{interacting objects} and treat all objects \textit{equally}, resulting in mismatched relation pairs. $\!$To this end, we propose an interaction-aware OVSGG framework INOVA. During pre-training, INOVA employs an interaction-aware target generation strategy to distinguish interacting objects from non-interacting ones. In SFT, INOVA devises an interaction-guided query selection tactic to prioritize interacting objects during bipartite graph matching. $\!$Besides, INOVA is equipped with an interaction-consistent knowledge distillation to enhance the robustness by pushing interacting object pairs away from the background. Extensive experiments on two benchmarks (VG and GQA) show that INOVA achieves state-of-the-art performance, demonstrating the potential of interaction-aware mechanisms for real-world applications.
\end{abstract}

\section{Introduction}
Scene graph generation~\cite{xu2017scene} (SGG) aims to map an image into a structured semantic representation, where objects are expressed as nodes and their relationships are as edges within the graph. Recently, with the burgeoning of large-scale models, \eg, vision-language models (VLMs) and multimodal large language models (MLLMs), open vocabulary SGG~\cite{he2022towards,li2024pixels,chen2024expanding} (OVSGG) has emerged as a promising area. It pushes beyond predefined categories to support the recognition and generation of novel objects and relationships, holding great potential for real-world applications.

\begin{figure}
    \centering
    \includegraphics[width=1\linewidth]{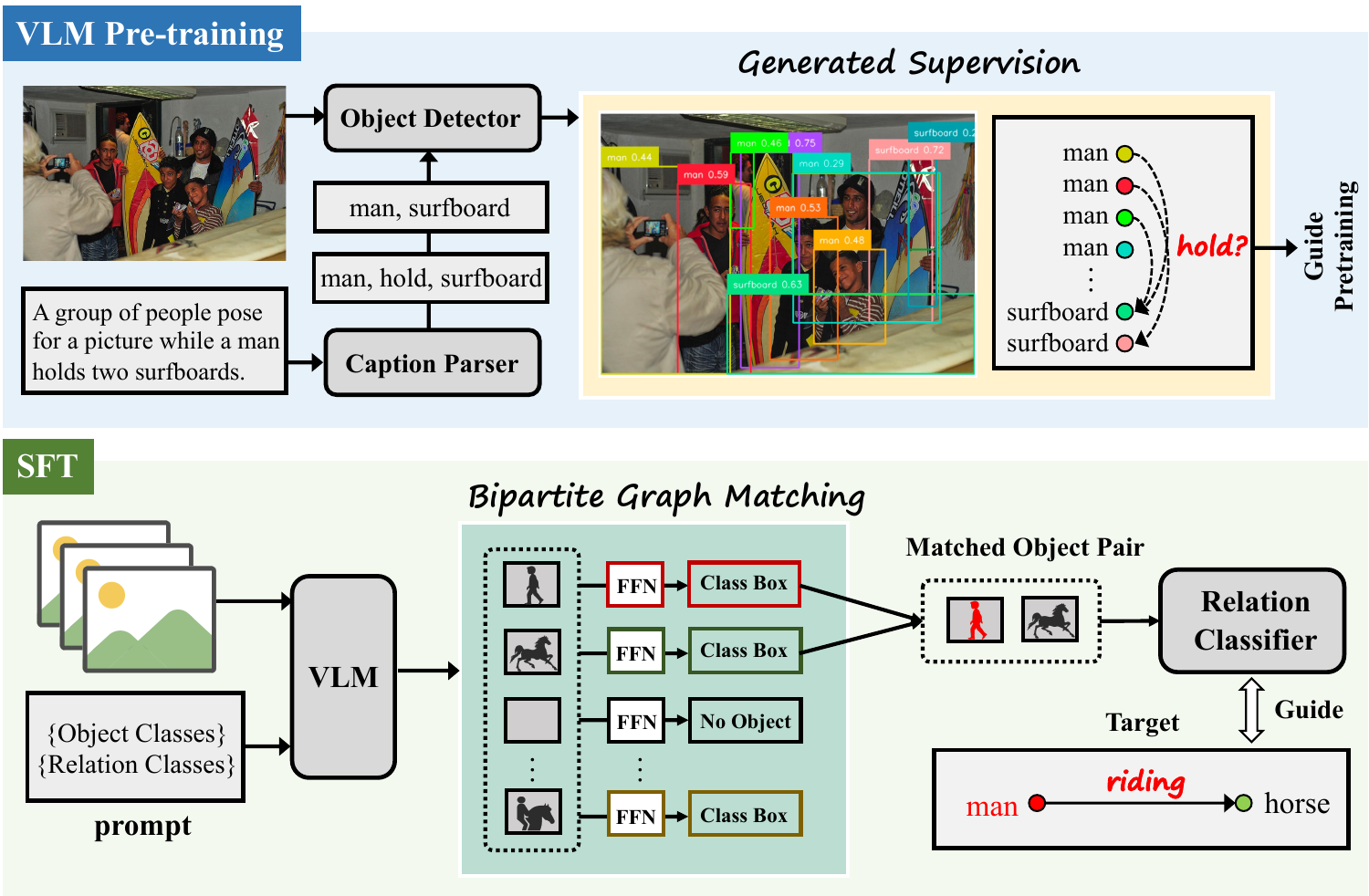}
    \vspace{-2em}
    \caption{ Overview of the OVSGG framework challenges.
1) VLM Pre-training, using solely entity categories for object detection causes ambiguity in associating object pairs (\eg, identifying the correct ``\texttt{man}-\texttt{surfboard}'' for the ``\texttt{hold}'').
2) SFT, bipartite graph matching misaligns non-interacting objects (\eg, ``\texttt{man}\includegraphics[scale=0.4]{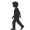}'') with interacting target ``\texttt{man}'' in $\langle$\texttt{man}, \texttt{riding}, \texttt{horse}$\rangle$.}
    \label{fig:intro}
\vspace{-1em}
\end{figure}

Generally, an end-to-end VLM-based\footnote{We primarily discuss VLM-based models here due to the high resource demands of MLLM-based approaches.} OVSGG pipeline consists of two stages: \textbf{VLM Pre-training} and \textbf{Supervised Fine-Tuning (SFT)}. The \textbf{former} involves pre-training a VLM on large-scale datasets to realize a visual-concept alignment by comparing the given caption and visual regions. Specifically, due to the lack of region-level information (\eg, bounding box annotations), recent work~\cite{he2022towards, zhang2023learning, chen2024expanding} adopts a weakly-supervised strategy to generate $\langle$\texttt{subject}, \texttt{predicate}, \texttt{object}$\rangle$ triplets with bounding boxes as pseudo supervisions. As displayed in Figure~\ref{fig:intro}(a), this approach extracts semantic graphs from image captions using SGG parsers~\cite{schuster2015generating}, then grounds objects in the graphs with pre-trained object detectors (\eg, Faster R-CNN~\cite{ren2015faster}, GLIP~\cite{li2022grounded} and Grounding DINO~\cite{liu2023grounding}). The \textbf{latter} stage refines the model's performance on task-specific objectives by leveraging high-quality annotations. Concretely, it fine-tunes part of VLM's parameters~\cite{chen2024expanding} or adapts prompt-tuning~\cite{he2022towards} on SGG dataset with fully-supervised triplet annotations. Leveraging these bounding box annotations, a DETR-like structure~\cite{carion2020end} with bipartite graph matching is typically used to align predicted entities with ground-truth labels. This stage further enhances the model's capability to recognize and generate precise scene graphs (\cf Figure~\ref{fig:intro}(b)).

Despite impressive, existing OVSGG methods often treat all objects \textit{equally}, ignoring the distinct characteristics of the \textbf{interacting objects}. By \textit{equally}, we mean the lack of differentiation between instances within the same category. For example, the \texttt{man} involved in a holding action and the \texttt{man} without any action are represented in an indistinguishable manner. It can lead to \textbf{\emph{mismatches in relation pairs}} during both pre-training and SFT stages, which induces the following drawbacks: \hypertarget{Q1}{\ding{182}} \textit{Bringing noisy supervision in pre-training}. As illustrated in Figure~\ref{fig:intro}(a), relying solely on entity categories (\eg, \texttt{man} and \texttt{surfboard}) to detect objects generates a large number of candidate pairs. This ambiguity makes it hard to associate relation (\eg, ``\texttt{hold}'') to the proper object pair (\eg, ``\texttt{man}-\texttt{surfboard}''). Using mismatched triplets (\eg, \texttt{man} in red and \texttt{surfboard} in pink) further exacerbates the confusion, hindering the training of robust SGG models. \myhyperlink{Q2}{\ding{183}} \textit{Leading mismatched bipartite graph during SFT}. In Figure~\ref{fig:intro}(b), a non-interacting ``\texttt{man}\includegraphics[scale=0.4]{figures/man1.png}'' can be mistakenly associated with \texttt{man} in the triplet annotation $\langle$\texttt{man}, \texttt{riding}, \texttt{horse}$\rangle$. However, the real target is another ``\texttt{man}~\scalebox{-1}[1]{\includegraphics[scale=0.4]{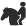}}'' engaged in riding. This mismatch further complicates the relation classification task, making it harder to predict correct interactions.

In this paper, we take a closer look at interacting objects in each stage, and propose the \underline{\textbf{IN}}teraction-aware \underline{\textbf{O}}pen-\underline{\textbf{V}}oc\underline{\textbf{A}}bulary SGG framework (\textbf{INOVA}).  INOVA follows a dual-encoder-single-decoder architecture~\cite{liu2023grounding}, comprising three key components: the visual and text encoders, the cross-modality decoder, and the entity and relation classifiers. During the VLM pre-training stage, INOVA introduces an \textbf{interaction-aware target generation} strategy that employs bidirectional interaction prompts to guide the grounding of interacting object pairs. These prompts incorporate interaction tokens that capture contextual dependencies and relational semantics, enabling the model to distinguish interacting objects from non-interacting ones through the attention mechanism~\cite{vaswani2017attention}. For the SFT stage, we devise a two-step \textbf{interaction-guided query selection} mechanism to prioritize interacting objects and incorporate relational context into the query selection process. This mechanism mitigates the interference of inactive objects and reduces mismatches in bipartite graph matching, ensuring robust relation prediction. Additionally, to distinguish interacting objects (engaged in both seen and unseen triplets) from the background and address the challenge of catastrophic knowledge forgetting~\cite{chen2024expanding} during SFT, we adopt an \textbf{interaction-consistent knowledge distillation} (KD). It utilizes a teacher model pre-trained on image-caption data to guide the student model in preserving both point-wise semantic alignment and inter-pair relational consistency. By explicitly modeling the relative dependencies between interaction-based and non-interaction pairs, it enhances the model's robustness in handling novel triplet combinations and background. 

To evaluate INOVA, we conducted comprehensive experiments on benchmark Visual Genome (VG)~\citep{krishna2017visual} and GQA~\citep{hudson2019gqa} datasets to validate its effectiveness in addressing the key challenges of OVSGG. In summary, our contributions are threefold:
\begin{itemize}[itemsep=0pt, parsep=0pt, topsep=0pt, partopsep=0pt]
\item We reveal key limitations in existing OVSGG frameworks, \ie, treating all objects \textit{equally}, which neglects the distinct characteristics of interacting objects and results in mismatched relation pairs.
\item We propose the INOVA framework that incorporates interaction-aware target generation, interaction-guided query selection, and interaction-consistent KD to pay attention to interacting objects, alleviating mismatched relation pairs and interference of irrelevant objects.
\item Extensive experiments on two prevalent SGG benchmarks demonstrate the effectiveness of INOVA.
\end{itemize}

\section{Related Work}


\textbf{OVSGG.}
This task bridges the gap between closed-set SGG and real-world requirements by leveraging VLMs or MLLMs to generalize beyond predefined categories~\cite{radford2021learning, liu2023grounding}. Current approaches fall into two main categories:
1) \textit{VLM-based Methods.} These approaches primarily rely on contrastive pre-training to align visual and textual embeddings. By comparing visual features of unseen objects or relations and their semantic counterparts in common semantics spaces, these models (\eg, CLIP~\cite{radford2021learning} and Grounding DINO~\cite{liu2023grounding}) enables zero-shot generalization. Recent advancements, such as He \etal~\cite{he2022towards}, explore visual-relation pre-training and prompt-based fine-tuning for OVSGG. Yu \etal~\cite{yu2023visually} leverage CLIP to align relational semantics in multimodal spaces, while Chen \etal~\cite{chen2024expanding} use a student-teacher framework to improve open-set relation prediction. Besides, other methods integrate category descriptions~\cite{li2024zero} or scene-level descriptions~\cite{chen2024scene} to enrich the semantic context and improve the discrimination among different relationships.
2) \textit{MLLM-based Methods.} These tactics extend the capabilities of VLMs by incorporating auto-regressive language models, predicting objects and relations in an open-ended manner. Specifically, they utilize the sequential prediction capabilities of MLLMs, \eg, BLIP~\cite{li2023blip} and LLaVA~\cite{liu2024visual}, to model scene graphs as structured sequences. For example, PGSG~\cite{li2024pixels} and OpenPSG~\cite{zhou2025openpsg} employ auto-regressive modeling to iteratively predict objects and relations, providing fine-grained relational reasoning for open-set triplets. ASMv2~\cite{wang2025all} builds on LLaVA~\cite{liu2024visual} with instruction fine-tuning, unifying text generation, object localization, and relation comprehension. Despite their power, MLLM-based methods typically require huge computing resources. In this paper, we focus on VLM-based methods and propose an interaction-aware framework that explicitly models object interactions and enhances generalization to novel categories.

\textbf{Weakly Supervised SGG.} This task aims to train models using language descriptions instead of fully annotated scene graphs. Existing works usually extract entities and relations from captions using language parsers~\cite{schuster2015generating}, then ground corresponding regions. Grounding methods include contrastive learning-based graph matching~\cite{shi2021simple}, semantic matching rules~\cite{zhong2021learning}, knowledge distillation from pre-trained VLMs~\cite{li2022integrating}, and aligning regions and words for scene graph supervision~\cite{zhang2023learning}. Recent large language model (LLM)-based approaches, \eg, LLM4SGG~\cite{kim2024llm4sgg} uses LLM’s reasoning capabilities to refine triplet extraction and alignment, mitigating semantic over-simplification. Similarly, GPT4SGG~\cite{chen2023gpt4sgg} synthesizes holistic and region-specific narratives, using the generative power of GPT-4~\cite{openai2023gpt} to capture both global context and local details. In this paper, we propose a simple and efficient method that only use LLM to generate counter-actions involved in bidirectional interaction prompts to improve interacting object detection accuracy.

\textbf{Knowledge Distillation (KD).}
This strategy trains a smaller ``student'' model to replicate the outputs of a larger ``teacher'' model, commonly used in open-vocabulary learning to transfer knowledge from VLMs. It encourages the student to mimic the teacher's enriched hidden space, enabling generalization from base to novel concepts. Prior work~\cite{gu2021open,zang2022open} explores KD in open vocabulary object detection by using L1/MSE loss to align the student detector's features with the teacher VLM's regional visual features. 
However, this hard alignment may fail to capture complex feature structures. Later work~\cite{NEURIPS2022_dabf6125} aligns the similarity of inter-embeddings, aiding in the acquisition of structured knowledge. Recent work extends to multi-scale level~\cite{wang2023object} or bags-of-region level~\cite{wu2023aligning}, contrasting with InfoNCE loss. This paper adopts an interaction-consistent KD that combines point-to-point concept retention and structure-aware interaction retention distillation, preserving teacher's knowledge and identifying novel relationships beyond backgrounds.

\section{Methodology}

\subsection{OVSGG Pipeline Review}
\textbf{Formulation.} Given an image $I$, SGG aims to construct a structured semantic graph $ \mathcal{G} = (\mathcal{V}, \mathcal{E}) $.  Each node $ v_i \in \mathcal{V} $ is defined by its bounding box (bbox) and category, while each edge $ e_{ij} \in \mathcal{E} $ represents the relationship between $ v_i $ and $ v_j $. In \textbf{open-vocabulary settings}, the label set $ \mathcal{C} $ for nodes and edges is divided into \textit{base classes} $ \mathcal{C}_B $ and \textit{novel classes} $ \mathcal{C}_N $, such that $ \mathcal{C}_B \cup \mathcal{C}_N = \mathcal{C} $ and $ \mathcal{C}_B \cap \mathcal{C}_N = \emptyset $. $ \mathcal{C}_B $ contains seen classes during training, while $ \mathcal{C}_N $ includes unseen classes that the model is expected to generalize to during inference.

\subsubsection{Architecture}
\begin{figure*}
    \centering
    \includegraphics[width=1\linewidth]{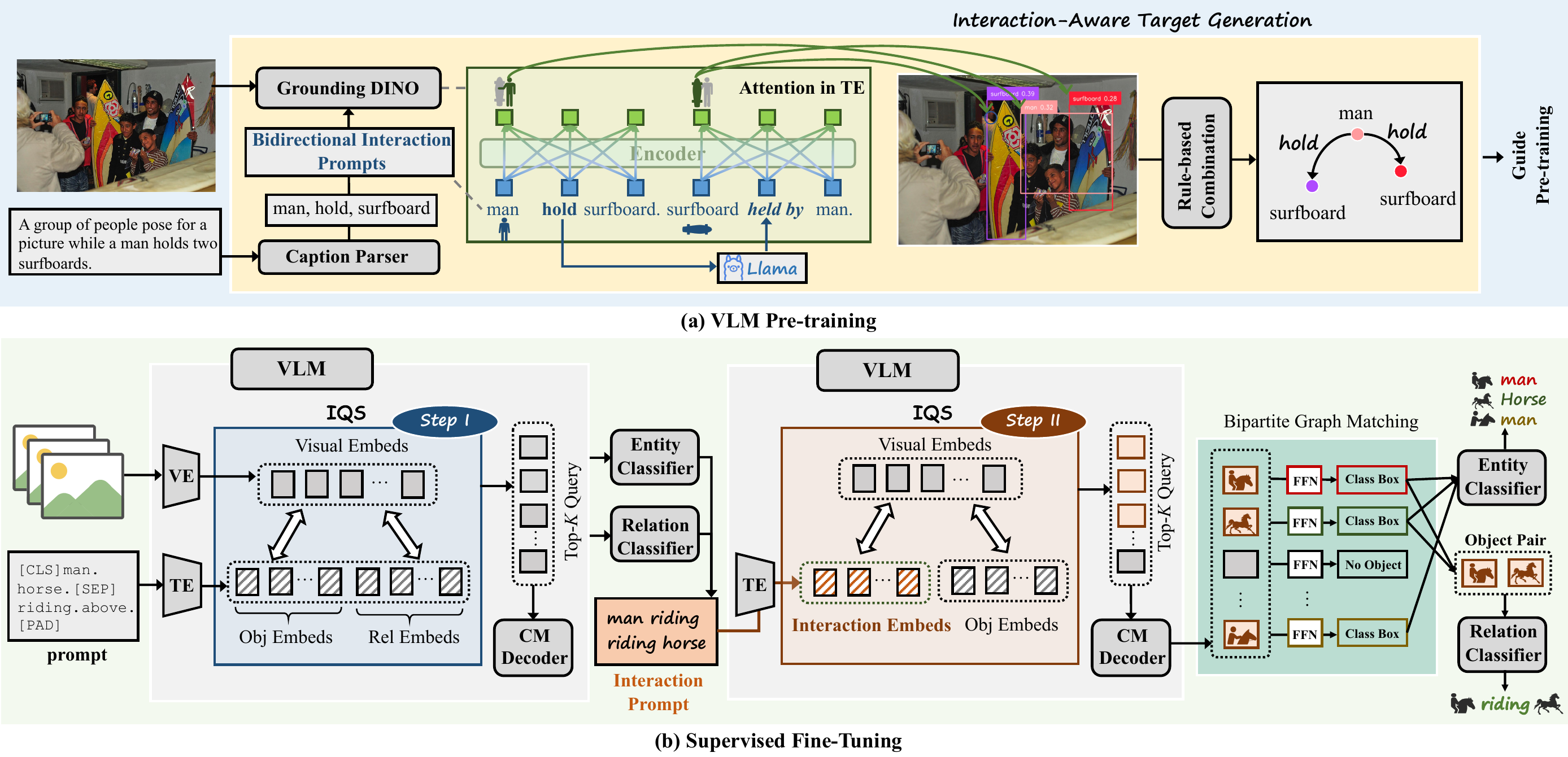}
    \put(-388, 68){\tiny{(Eq.~\ref{eq:step1})}}
    \put(-205, 68){\tiny{(Eq.~\ref{eq:step2})}}
    \vspace{-2em}
    \caption{Overview of INOVA for OVSGG. (a) VLM Pre-training: Interaction-aware target generation uses bidirectional interaction prompts and rule-based bounding box combinations to generate supervision, enriching object tokens with contextual interaction semantics. (b) SFT: A two-step interaction-guided query selection (IQS) prioritizes interacting objects and integrates relational context into object tokens, refining queries for the decoder. Bipartite graph matching aligns predictions with ground-truth for entity and relation classification.}
    \label{fig:framework}
\end{figure*}

As illustrated in Figure~\ref{fig:framework}(b), an end-to-end OVSGG framework~\cite{chen2024expanding} typically follows a dual-encoder-single-decoder architecture~\cite{liu2023grounding}, involving three main components: the visual and text encoders, the cross-modality decoder, the entity and relation classifiers.

\textbf{Visual and Text Encoders.}
The visual encoder (VE) extracts multi-scale visual features $\mathbf{V} \in \mathbb{R}^{N_v \times d}$ by the image backbone (\eg, Swin Transformer~\cite{liu2021swin}). For the text encoder (TE), input prompts are constructed by concatenating all predefined object and relation categories into a single sequence, \eg, ``\texttt{[CLS]} \texttt{man}. \texttt{horse}. \texttt{[SEP]} \texttt{riding}. \texttt{above}. \texttt{[PAD]}'', following~\cite{chen2024expanding}. Using this prompt, TE extracts object features $\mathbf{T}_o \in \mathbb{R}^{N_o \times d}$ and relation features $\mathbf{T}_r \in \mathbb{R}^{N_r \times d}$ using a pre-trained language model (\eg, BERT~\cite{kenton2019bert}). Here, $N_v$, $N_o$, and $N_r$ denote the numbers of image, object, and relation tokens, respectively. $d$ is the feature dimension.

\textbf{Cross-Modality (CM) Decoder.}  
It refines the representations of $K$ object queries $\{\mathbf{q}_i\}_{i=1}^K$ through a series of operations, including a self-attention layer, an image cross-attention layer for visual features, and a cross-attention layer for text features derived from prompts~\cite{liu2023grounding}. These refined queries are then passed through a feed-forward network (FFN) to predict object bbox coordinates. Following~\cite{chen2024expanding, shit2022relationformer}, a global relation query $\mathbf{q}_{rel}$ is introduced to capture spatial and semantic dependencies among objects in the image, complementing the local interactions represented by the object queries.

\textbf{Entity and Relation Classifiers.}
The entity/relation classifier compares node/edge features with text features of object/relation classes in a shared semantic space for open-vocabulary recognition. $\!$Concretely, node features $\{\mathbf{e}_o\}$ are from refined object queries and edge features $\{\mathbf{e}_{ij}\}$ are constructed by combining paired object features to capture subject-object interactions. $\!$To model interactions effectively, VS~\cite{zhang2023learning} constructs edge features by computing the differences and sums of object features. In contrast, we follow~\cite{chen2024expanding,shit2022relationformer} to concatenate a global relation embedding $\mathbf{e}_{rln}$ (refined representation of relation query) with pairwise object embeddings. The concatenated features are through a two-layer MLP to capture holistic dependencies and interactions for $\mathbf{e}_{ij}$.

\subsubsection{Training Process}
\label{sec:method_train}
\textbf{Bipartite Graph Matching.}
During training, it matches object queries with ground-truth (GT) annotations by minimizing a cost function based on semantic similarity and spatial alignment~\cite{carion2020end}. Matched queries are used for entity classification and linked to the matched GT's subordinate triplet, serving as input for edge representations.

\textbf{Training Objectives.}
Following~\cite{chen2024expanding}, there are three training losses: 1) \textit{Bbox Regression Loss}: Combines L1 $\mathcal{L}_{reg}$ and GIoU loss $\mathcal{L}_{giou}$~\cite{rezatofighi2019generalized} to ensure accurate object localization with precise positions and bounding box overlaps. 2) \textit{Entity Classification Loss}: Applies Focal Loss~\cite{lin2017focal} $\mathcal{L}_{obj}$ to alleviate imbalance-distribution issue by focusing on hard-to-classify and underrepresented object categories. 3) \textit{Relation Classification Loss}:
Uses binary cross-entropy (BCE) loss $\mathcal{L}_{rel}$ to align predicted relation scores with GT annotations.

\subsection{INOVA}
As illustrated in Figure~\ref{fig:framework}, INOVA follows a two-stage training process, incorporating interaction-aware target generation during pre-training and interaction-guided query selection in SFT to alleviate mismatches caused by uniform treatment of objects in each stage. Besides, an interaction-consistent KD further enhances the model's ability to distinguish interaction-based pairs from background noises.

\subsubsection{Interaction-Aware Target Generation}
\label{sec:itg}

To effectively identify interacting objects in weakly annotated data during pre-training, $\!$we devise an interaction-aware target generation tactic that uses bidirectional triplets rather than relying on a direct combination of all entity classes (\eg, ``\texttt{man}. \texttt{surfboard}.'') for object detection.

To be specific, after the semantic graph parsing process, we employ Grounding DINO~\cite{liu2023grounding} as the object detector and design \textbf{bidirectional interaction prompt} to guide the object localization. The bidirectional interaction prompt is constructed by combining two perspectives for each interaction triplet: one reflecting the action from the subject’s viewpoint (\eg, ``\texttt{man hold surfboard}'') and another from the object’s perspective (\eg, \texttt{surfboard held by man}''). The former is directly derived from the components of the interaction triplet, while the latter converse the subject and object with a \textit{counter-action} (\eg, ``\texttt{held by}'') generated by an LLM (\eg, Llama2~\cite{touvron2023llama})\footnote{The generation process of counter-action is in the Appendix C.\label{footnote:counterrel_gen}}. The dual-perspective construction process brings two key advantages: 1) \textit{Modeling Context Information}: Through the attention mechanism in the text encoder of Grounding DINO, the bidirectional interaction prompt integrates contextual interaction information into object tokens. As shown in Figure~\ref{fig:framework}(a), the attention mechanism enables the token ``\texttt{man}'' to absorb relevant interaction semantics, such as ``\texttt{hold surfboard}'', ensuring that the grounded object ``\texttt{man}'' is correctly aligned with its interaction context.
2) \textit{Enhancing Object Role Awareness}: By reversing operation, the object (\eg, ``\texttt{surfboard}'') of given triplet becomes the syntactic subject of the whole sentence (\eg, ``\texttt{surfboard held by man}''). As the central of the rephrased sentence, the syntactic subject receives heightened attention, improving its accuracy in localization. 

Furthermore, inspired by~\cite{li2022integrating,kim2024llm4sgg}, we adopt a \textit{rule-based combination} that combines overlapping subject and object bounding boxes to form reliable triplet supervision by Intersection over Union (IoU) score.

\subsubsection{Interaction-Guided Query Selection}
\label{sec:iqs}
During SFT, we introduce a two-step selection strategy for query initialization and refinement to prioritize interacting objects, mitigating the bipartite graph mismatched problem by reducing non-interacting candidates.

\textbf{Step I.}  
This step aims to directly identify the most relevant visual tokens that are likely to participate in object interactions. Intuitively, the visual features of interacting objects should exhibit strong correlations with both object and relation semantics. To achieve this, for each visual token $\mathbf{v}_i \in \mathbf{V}_v$, a relevance score $s_i$ is computed by combining its maximum similarity with object and relation class tokens:
\begin{equation}
\small
\label{eq:step1}
    s_i = \big( \max(\mathbf{v}_i \mathbf{T}_{o}^\top) \big)^\gamma \cdot \big( \max(\mathbf{v}_i \mathbf{T}_{r}^\top) \big)^{1-\gamma},
\end{equation}
where $\max(\mathbf{v}_i \mathbf{T}_{o}^\top)$ computes the maximum similarity between the visual token $\mathbf{v}_i$ and all object class tokens in $\mathbf{T}_{o}$, while $\max(\mathbf{v}_i \mathbf{T}_{r}^\top)$ computes the maximum similarity between $\mathbf{v}_i$ and all relation class tokens in $\mathbf{T}_{r}$. The parameter $\gamma\in[0, 1]$ balances their contributions.

Based on the relevance scores, the top $K$ query indices, denoted as $\mathcal{I}_{K}$, are selected by the following procedure:
\begin{equation}
\small
    \mathcal{I}_{K} = \text{Top}_{K} ( \{ s_i \mid i = 1, 2, \dots, N_v \} ).
\end{equation}
The visual features and the position embedding~\cite{liu2023grounding} corresponding to the selected indices $\mathcal{I}_{K}$ are used to initialize queries for further decoding operations.

\textbf{Step II.}  
Nevertheless, the object and relation tokens are encoded individually in Step I, which limits capturing interaction semantics and distinguishing among objects. To this end, Step II explicitly models interaction semantics by integrating relational context into the object tokens. 
Specifically, after the initial forward pass, the model predicts a set of visual relation triplets $\langle$\texttt{subject}, \texttt{predicate}, \texttt{object}$\rangle$. These triplets are decomposed into interaction pairs $\langle$\texttt{subject}, \texttt{predicate}$\rangle$ and $\langle$\texttt{predicate}, \texttt{object}$\rangle$, which serve as \textbf{interaction prompts}. These prompts are encoded through the TE of VLM to get interaction tokens embeddings $\mathbf{T}_{in}$. The decomposition process has dual advantages: First, by leveraging interaction prompts, the TE's attention mechanism integrates interaction information into the object tokens, enabling the model to capture contextual dependencies and enhance its understanding of relationships. For instance, the token ``\texttt{man}'' can incorporate the semantic meaning of the interaction ``\texttt{riding}'' to obtain ``\texttt{man}~\scalebox{-1}[1]{\includegraphics[scale=0.4]{figures/man2.png}}'' in Figure~\ref{fig:framework}(b). Second, decomposing triplets into pairs avoids direct interference between object tokens, effectively preserving their unique characteristics. As illustrated in Figure~\ref{fig:framework}(b), ``\texttt{man}~\scalebox{-1}[1]{\includegraphics[scale=0.4]{figures/man2.png}}'' and ``\texttt{horse}~{\includegraphics[scale=0.04]{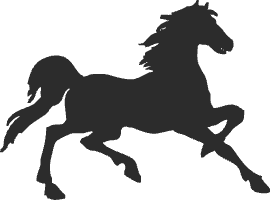}}'' are independently processed, preventing unnecessary dependencies across unrelated categories and maintaining the individual semantics of each object.

\textit{Interaction Query Selection.} For each visual token $\mathbf{v}_i$, the interaction relevance score $s_i^{in}$ is calculated by measuring the maximum similarity with interaction tokens:
\begin{equation}
\small
\label{eq:step2}
    s_i^{in} = \max (\mathbf{v}_i \mathbf{T}_{in}^\top).
\end{equation}
 The query indices set prioritizes the top $L$ tokens with the highest interaction relevance:
\begin{equation}
\small
    \mathcal{I}_L^{in} = \text{Top}_L ( \{ s_i^{in} \mid i = 1, 2, \dots, N_v \} ).
\end{equation}
\textit{Missing Query Selection.} However, relying solely on interaction relevance may fail to identify objects absent from the initially predicted triplets yet crucial for comprehensive scene understanding. To address this, the object relevance score $s_i^{o}$ is computed similarly, but using object tokens $\mathbf{T}_{o}$. The remaining $K-L$ query indices are selected based on object relevance, excluding those already chosen:
\begin{equation}
\small
    \mathcal{I}_{K-L}^{o} = \text{Top}_{K-L} (\{ s_i^o \mid i \notin \mathcal{I}_L^{in}, i = 1, 2, \dots, N_v \} ).
\end{equation}
The final query indices set combines these two subsets:
\begin{equation}
\mathcal{I}_K = \mathcal{I}_L^{in} \cup \mathcal{I}_{K-L}^{o}. \end{equation}
Combining Step I and Step II, the query selection achieves both interaction relevance and comprehensive integration of relational context. Step I identifies interaction-relevant tokens by balancing object and relation semantics, while Step II refines the representation by embedding relational context into object tokens through interaction prompts. This two-step strategy effectively reduces non-interacting candidates and mitigates mismatches in the bipartite graph. For ease of understanding, the pseudo-codes is left in Appendix D.


\subsubsection{Interaction-Consistent KD}  
\label{sec:ickd}
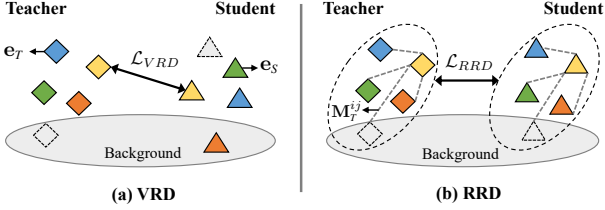
\begin{figure}
    \centering
    \includegraphics[width=1.0\linewidth]{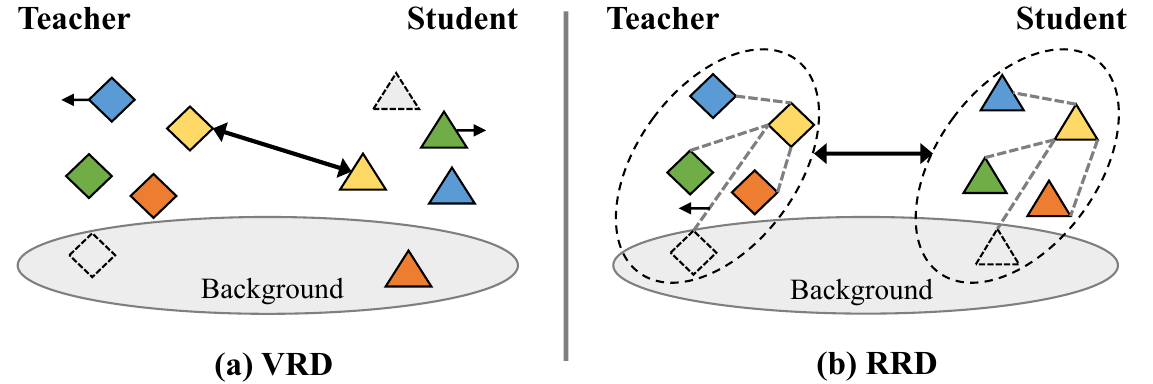}
    \put(-184, 53){\scalebox{0.7}{$\mathcal{L}_{VRD}$}}
    \put(-65, 52){\scalebox{0.7}{$\mathcal{L}_{RRD}$}}  
    \put(-232, 56){\scalebox{0.8}{$\mathbf{e}_{{\textit{\tiny{T}}}}$}}  
    \put(-135, 51){\scalebox{0.8}{$\mathbf{e}_{\text{\textit{\tiny{S}}}}$}} 
    \put(-108, 33){\scalebox{0.65}{$\mathbf{M}_{\text{\textit{\tiny{T}}}}^{ij}$}} 
    \vspace{-2em}
    \caption{Illustration of interaction-consistent KD.}
    \label{fig:kd}
    \vspace{-1em}
\end{figure}
Beyond the localization and classification objectives mentioned in Sec.~\ref{sec:method_train}, we adopt interaction-consistent knowledge distillation to enhance the model's ability to distinguish interacting pairs from background pairs and address catastrophic forgetting of learned relational semantics mentioned in~\cite{chen2024expanding}. Specifically, it leverages the VLM pre-trained in the first stage as the teacher model. The student network is designed as a pseudo-siamese structure of the teacher model, initialized with the teacher's parameters.  

Interaction-consistent KD combines visual-concept retention distillation and relative-interaction retention distillation to align the student model with the teacher's semantic space while maintaining inter-pair relational consistency. The entire loss function contains two complementary objectives:

\textit{Visual-concept Retention Distillation (VRD)}: As proposed in~\cite{chen2024expanding}, this objective ensures that the student's edge features remain point-wise consistent with the teacher's semantic space for negative samples, thereby preserving semantic alignment. The loss is defined as:
\begin{equation}
\small
\mathcal{L}_{{VRD}} = \frac{1}{|\mathcal{N}|} \sum_{\mathbf{e} \in \mathcal{N}} \| \mathbf{e}_{\text{\textit{\tiny{S}}}} - \mathbf{e}_{\text{\textit{\tiny{T}}}} \|_1,
\label{eq:visual_concept_kd}
\end{equation}
where $\mathbf{e}_{\text{\textit{\tiny{S}}}}$ and $\mathbf{e}_{\text{\textit{\tiny{T}}}}$ denote the edge features of the student and teacher models, and $\mathcal{N}$ is the set of negative samples.

\textit{Relative-interaction Retention Distillation (RRD)}: While VRD effectively preserves point-wise semantic consistency, it fails to ensure the relative relationships between triplets, \ie, distinguishing interaction pairs from backgrounds (\cf Figure~\ref{fig:kd}(a)). $\!$RRD explicitly models inter-pair relativity~\cite{NEURIPS2022_dabf6125} by aligning the structure similarity of triplet embeddings between the teacher and student models. The structure similarity matrices for the teacher and student models, $\mathbf{M}_{\text{\textit{\tiny{T}}}}$ and $\mathbf{M}_{\text{\textit{\tiny{S}}}}$, are normalized by L2 norm:
\begin{equation}
\small
\mathbf{M}_{\text{\textit{\tiny{T}}}}^{ij} = \frac{ \mathbf{e}_{\text{\textit{\tiny{T}}}}^i \cdot \mathbf{e}_{\text{\textit{\tiny{T}}}}^{j\top} }{\| \mathbf{e}_{\text{\textit{\tiny{T}}}}^i \cdot\mathbf{e}_{\text{\textit{\tiny{T}}}}^{j\top} \|_{2}}, \quad
\mathbf{M}_{\text{\textit{\tiny{S}}}}^{ij} = \frac{ \mathbf{e}_{\text{\textit{\tiny{S}}}}^i \cdot \mathbf{e}_{\text{\textit{\tiny{S}}}}^{j\top} }{\| \mathbf{e}_{\text{\textit{\tiny{S}}}}^i \cdot \mathbf{e}_{\text{\textit{\tiny{S}}}}^{j\top} \|_{2}}.
\label{eq:cosine_similarity}
\end{equation}
The RRD loss then aligns these similarity matrices by minimizing the Frobenius norm $\| \cdot \|_F$ between them:
\begin{equation}
\small
\mathcal{L}_{{RRD}} = \frac{1}{|\mathcal{N}|^2} \| \mathbf{M}_{\text{\textit{\tiny{S}}}} - \mathbf{M}_{\text{\textit{\tiny{T}}}} \|_F^2.
\label{eq:relative_interaction_kd}
\end{equation}
\textbf{Final Objectives:} Combine localization and classification losses with above complementary objectives to achieve point-wise semantic alignment and relational consistency:
\begin{equation}
\small
\mathcal{L} = \mathcal{L}_{reg} + \mathcal{L}_{giou}  + \mathcal{L}_{obj} + \mathcal{L}_{rel} + \beta_1 \mathcal{L}_{{VRD}} + \beta_2 \mathcal{L}_{{RRD}}.
\end{equation}
The weights $\beta_1$ and $\beta_2$ control the relative importance of semantic alignment and relational consistency.

\section{Experiments}
\subsection{Experiment setup}
\textbf{Datasets.} We evaluated INOVA on two SGG benchmarks: 1) \textbf{VG}~\cite{krishna2017visual} contains annotations for 150 object categories and 50 relation categories across 108,777 images. Following standard setup~\cite{xu2017scene}, 70\% of the images are used for training, 5,000 for validation, and the remaining for testing. For a fair comparison, we excluded images overlapping with the pre-training dataset of Grounding DINO~\cite{liu2023grounding}, retaining 14,700 test images as in~\cite{zhang2023learning}. $\!$2) \textbf{GQA}~\cite{hudson2019gqa} uses the GQA200 split~\cite{dong2022stacked, sudhakaran2023vision}, including 200 object categories and 100 predicate categories. We randomly sampled 70\% of the object and predicate categories as the base, and more details can be found in the Appendix A.

\textbf{Metrics.} We conducted experiments under the challenging Scene Graph Detection (\textbf{SGDET}) protocol~\cite{xu2017scene,krishna2017visual}, which requires detecting objects and identifying relationships between object pairs without GT object labels or bounding boxes. We reported: 1) \textbf{Recall@K} (\textbf{R@K}): The proportion of ground-truth triplets correctly predicted within the top-K confident predictions.
2) \textbf{Mean R@K} (\textbf{mR@K}): The average R@K across all categories.
\begin{table*}[!t]
    \small
    \centering
    \caption{Experimental results of OvR-SGG setting on VG~\cite{krishna2017visual} test set.}   
    \setlength\tabcolsep{7.5pt}
    \scalebox{0.94}{
    \begin{tabular}{|rl||c|ccc|ccc|} 
        \hline
        \thickhline
        \rowcolor{mygray}
        & & & \multicolumn{3}{c|}{Base+Novel (Relation)} & 
        \multicolumn{3}{c|}{Novel (Relation)}  \\
        \rowcolor{mygray}
        \multicolumn{2}{|c||}{\multirow{-2}[0]{*}{Method}} & \multirow{-2}[0]{*}{Backbone} & R@20  & R@50  & R@100  & R@20 & R@50 & R@100  \\ 
        \hline
        \hline
        IMP~\cite{xu2017scene}&$_\text{CVPR'17}$ & - & - & 12.56    &  14.65 & - & 0.00 & 0.00 \\
        MOTIFS~\cite{zellers2018neural}&$_\text{CVPR'18}$& - & - & 15.41 & 16.96 & - & 0.00& 0.00 \\
        VCTREE~\cite{tang2019learning}&$_\text{CVPR'19}$& - & - & 15.61 & 17.26 & - & 0.00 & 0.00  \\
        TDE~\cite{tang2020unbiased}&$_\text{CVPR'20}$ & - & - & 15.50 & 17.37 & - &0.00 & 0.00 \\ 
        \hline
        $\text{VS}^3$~\cite{zhang2023learning}&$_\text{CVPR'23}$& \multirow{4}[0]{*}{Swin-T} & - &15.60 & 17.30 & - & 0.00 & 0.00 \\ 
        OvSGTR~\cite{chen2024expanding}  &$_\text{ECCV'24}$& & - & 20.46 & 23.86 & - & 13.45 & 16.19 \\ 
        RAHP~\cite{liu2025relation}  &$_\text{AAAI'25}$& & - & 20.50   & 25.74 & - & 15.59  & 19.92  \\ 

        \textbf{INOVA} (\textbf{Ours}) & &  & \textbf{17.49} & \textbf{23.22} & \textbf{27.40} & \textbf{12.90} & \textbf{17.89} & \textbf{21.70} \\
        \hline  OvSGTR~\cite{chen2024expanding}  &$_\text{ECCV'24}$& \multirow{2}[0]{*}{Swin-B} & - & 22.89   & 26.65 & - & 16.39  & 19.72  \\ 
        \textbf{INOVA} (\textbf{Ours}) & &  &
        \textbf{18.77} & \textbf{24.81} &
        \textbf{29.28} & \textbf{14.72} & \textbf{20.04} & \textbf{24.66}\\
        \thickhline
    \end{tabular}
    }
    \label{tab:ovr}
    \vspace{-1.5em}
\end{table*}

\begin{table*}[!t]
    \small
    \centering
    \caption{Experimental results of OvD+R-SGG setting on VG~\cite{krishna2017visual} test set.}    
    \setlength\tabcolsep{2.5pt}
    \scalebox{0.94}{
    \begin{tabular}{|rl||c|ccc|ccc|ccc|} 
        \hline
        \thickhline
        \rowcolor{mygray}
        & & & \multicolumn{3}{c|}{Joint Base+Novel} & 
        \multicolumn{3}{c|}{Novel (Obj)} & 
        \multicolumn{3}{c|}{Novel (Rel)} \\
        \rowcolor{mygray}
        \multicolumn{2}{|c||}{\multirow{-2}[0]{*}{Method}}  & \multirow{-2}[0]{*}{Backbone} & R@20  & R@50  & R@100  & R@20 & R@50 & R@100  & R@20 & R@50 & R@100 \\ 
        \hline
        \hline
        IMP~\cite{xu2017scene}&$_\text{CVPR'17}$ & - & - &   0.77  &  0.94 & - & 0.00 & 0.00 & - & 0.00 & 0.00 \\
        MOTIFS~\cite{zellers2018neural}&$_\text{CVPR'18}$ & - & - & 1.00 & 1.12 & - & 0.00 & 0.00 & - & 0.00 & 0.00 \\
        VCTREE~\cite{tang2019learning}&$_\text{CVPR'19}$ & - & - & 1.04 & 1.17 & - & 0.00 & 0.00 & - & 0.00 & 0.00 \\
        TDE~\cite{tang2020unbiased}&$_\text{CVPR'20}$ & - & - &  1.00 & 1.15 & - &0.00 & 0.00 & - & 0.00 & 0.00 \\ 
        \hline
        $\text{VS}^3$~\cite{zhang2023learning}&$_\text{CVPR'23}$& \multirow{3}[0]{*}{Swin-T} & - &
          5.88 & 7.20 & - & 0.00 & 0.00 & - & 0.00 & 0.00 \\ 
        \text{OvSGTR}~\cite{chen2024expanding}&$_\text{ECCV'24}$& & 10.02 & 13.50 & 16.37 & 10.56 & 14.32 & 17.48 & 7.09 & 9.19 & 11.18 \\ 
        \textbf{INOVA} (\textbf{Ours}) & & & \textbf{12.61}  & \textbf{17.43} & \textbf{21.27} & \textbf{12.48} & \textbf{17.16} & \textbf{21.10} & \textbf{11.38} & \textbf{15.90} & \textbf{19.46} \\  
        \hline\text{OvSGTR}~\cite{chen2024expanding} &$_\text{ECCV'24}$&\multirow{2}[0]{*}{Swin-B} & 12.37 & 17.14   & 21.03 & 12.63 & 17.58  & 21.70  & 10.56 & 14.62 & 18.22 \\ 
        
        \textbf{INOVA} (\textbf{Ours}) & & &  \textbf{13.50} & \textbf{18.88} & \textbf{23.19} & \textbf{13.46} & \textbf{18.84} & \textbf{23.29} & \textbf{12.37} & \textbf{17.50} & \textbf{21.73} \\
        \thickhline
    \end{tabular}
    }
    \label{tab:ovdr}
    \vspace{-1em}
\end{table*}

\textbf{Implementation Details.}
Due to space constraints, detailed implementation is provided in the Appendix A.
\subsection{Comparison with State-of-the-Art Methods}
\textbf{Setting.} Following~\cite{chen2024expanding}, we compared our INOVA with existing SOTA methods, \ie, \textbf{VS}~\cite{zhang2023learning}, \textbf{OvSGTR}~\cite{chen2024expanding}, and \textbf{RAHP}~\cite{liu2025relation} under two OVSGG settings: 1) \textbf{OvR-SGG}: Evaluates generalization to unseen relations while retaining original object categories. Fifteen of 50 relation categories in VG150 are removed during training, with performance measured on ``Base+Novel (Relation)'' and ``Novel (Relation)''.
2) \textbf{OvD+R-SGG}: Assesses handling of unseen objects and relations simultaneously. Both novel objects and relations are excluded during training, evaluated on ``Joint Base+Novel'', ``Novel (Object)'', and ``Novel (Relation)''. 

\textbf{Results.}
We conducted quantitative experiments on the VG dataset~\cite{krishna2017visual} in both the OvR-SGG and OvD+R-SGG setups, with results presented in Table~\ref{tab:ovr} and Table~\ref{tab:ovdr}, respectively. Notably, INOVA consistently outperforms the latest state-of-the-art methods across all metrics. In the OvR-SGG setup, INOVA surpasses the RAHP (Swin-T) by \textbf{+1.78}\% R@100 within the novel relation categories, demonstrating superior generalization and reduced overfitting. With the Swin-B backbone, INOVA achieves R@100 over OvSGTR across both base and novel relations, and \textbf{+4.94}\% R@100 in novel relations alone, further emphasizing its robustness. In the more challenging OvD+R-SGG scenario, INOVA continues to outperform the competition. Specifically, on the joint base and novel classes, INOVA gains \textbf{+4.90}\% and \textbf{+2.16}\% R@100 over OvSGTR with the Swin-T and Swin-B backbones, respectively. These results validate INOVA's superior performance and robust generalization across both relation and object domains.

\subsection{Diagnostic Experiment}

\begin{table*}[!t]
    \small
    \centering
    \caption{Analysis of key components on OvD+R-SGG setting of VG150~\cite{krishna2017visual} test set. \textbf{ITG}, \textbf{IQS}, and \textbf{RRD} stand for Interaction-aware Target Generation, Interaction-guided Query Selection, and Relative-interaction Retention Distillation in interaction-consistent knowledge distillation, respectively. The general OVSGG pipeline with visual-concept retention distillation as the baseline.}
    \setlength\tabcolsep{7.5pt}
    \scalebox{0.94}{
    \begin{tabular}{|ccc||ccc|ccc|ccc|} 
        \hline
        \thickhline
        \rowcolor{mygray}
        \multicolumn{3}{|c||}{Components}  & \multicolumn{3}{c|}{Joint Base+Novel} & 
        \multicolumn{3}{c|}{Novel (Obj)} & 
        \multicolumn{3}{c|}{Novel (Rel)} \\
        \rowcolor{mygray}
         ITG & IQS & RRD & R@20  & R@50  & R@100  & R@20 & R@50 & R@100  & R@20 & R@50 & R@100 \\ 
        \hline
        \hline
        &  &  & 10.02 & 13.50 & 16.37 & 10.56 & 14.32 & 17.48 & 7.09 & 9.19 & 11.18 \\
        & & \usym{1F5F8} & 11.43  & 15.67 & 19.20 & 11.57 & 15.65 & 19.32 & 10.07 & 14.00 & 17.32 \\
        & \usym{1F5F8} & & 11.37 & 15.71  & 19.37  & 11.43 & 15.80 & 19.61 & 9.84 & 13.92 & 17.38  \\
        \usym{1F5F8} &  &  &  11.92 & 16.67 & 20.31 & 11.75 & 16.51 & 20.16 & 10.72 & 15.10 & 18.52 \\
        & \usym{1F5F8} & \usym{1F5F8} &11.84  & 16.17 & 19.55 & 11.36 & 16.09 & 19.65 & 10.73 & 14.40 & 17.83  \\
        \usym{1F5F8} & & \usym{1F5F8} & 12.27  & 17.11 & 20.81 & 12.16 & 17.03 & 20.80 & 11.04 & 15.60 & 19.01 \\
        \usym{1F5F8} & \usym{1F5F8} &  & 12.42  & 17.22 & 21.10 & 12.29 & 17.08 & 20.99 & 11.16 & 15.51 & 19.16 \\
        \usym{1F5F8} & \usym{1F5F8} & \usym{1F5F8} &\textbf{12.61}  & \textbf{17.43} & \textbf{21.27} & \textbf{12.48} & \textbf{17.16} & \textbf{21.10} & \textbf{11.38}& \textbf{15.90} & \textbf{19.46} \\
        \thickhline
    \end{tabular}
    }
    \vspace{-1.5em}
    \label{tab:abla}
\end{table*}
To ensure a comprehensive evaluation, we performed a series of ablation studies on the VG dataset~\cite{krishna2017visual} in the challenging OvD+R-SGG scenario.

\textbf{Key Components Analysis.} 
The results are summarized in Table~\ref{tab:abla}, with the first row representing the baseline OVSGG pipeline with \textit{Visual-concept Retention Distillation} proposed in~\cite{chen2024expanding}. From this analysis, four key conclusions can be drawn: \textbf{First}, incorporating \textit{Interaction-aware Target Generation} (ITG) leads to consistent improvements across all metrics, including a \textbf{3.94}\% R@100 gain on the joint base and novel classes compared to the baseline. This demonstrates that ITG effectively improves performance by considering interaction contexts in supervision generation. \textbf{Second}, introducing \textit{Interaction-guided Query Selection} (IQS) further refines the query selection process. By prioritizing interacting objects and minimizing mismatched assignments, IQS achieves notable improvements, such as \textbf{3.00}\% R@100 gains, highlighting its ability to enhance precision by focusing on interacting object pairs. \textbf{Third}, leveraging \textit{Relative-interaction Retention Distillation} (RRD) ensures relational consistency during training, resulting in significant performance boosts. RRD contributes \textbf{2.83}\% R@100 gains, improving the model's ability to handle novel classes effectively. \textbf{Fourth}, the integration of all three components (\ie, ITG, IQS, and RRD) yields the best overall performance, with \textbf{1.92}\%$\sim$\textbf{8.28}\% improvements across all evaluation metrics. However, the improvement is less pronounced than expected, since each strategy prioritizes interacting objects, which may lead to diminishing returns by progressively reducing non-interacting objects. Despite this, the combined results still demonstrates enhanced relational understanding and serve as a valuable tool for improving performance in complex scenarios.

\begin{figure}[!t]
    \centering
    \includegraphics[width=1\linewidth]{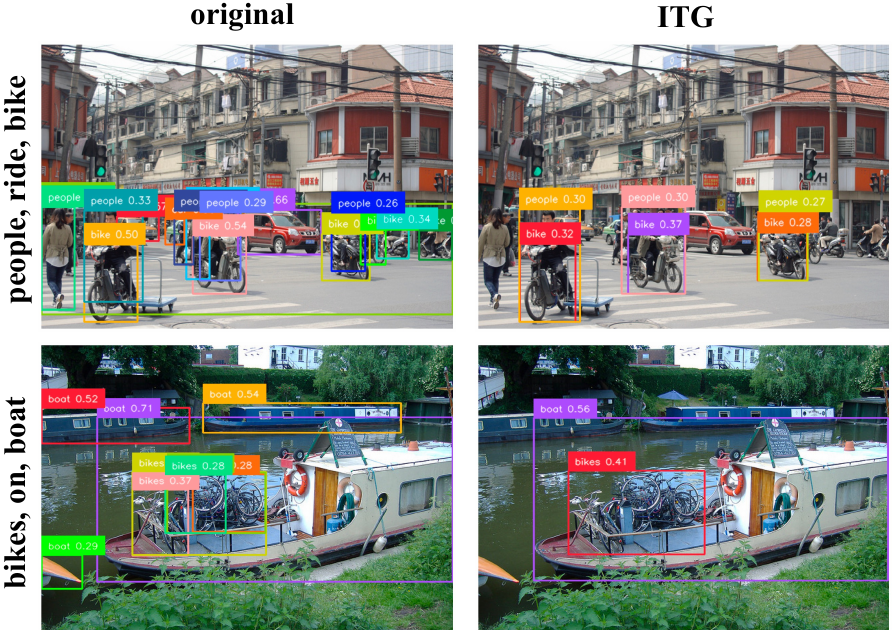}
    \vspace{-2.0em}
    \caption{Interaction-aware target generation.}
    \label{fig:itg}
    \vspace{-1.5em}
\end{figure}

\begin{table*}[t]
    \small
    \centering
    \caption{Comparison with pre-training methods. All models are \textbf{pre-trained} on image-caption data and tested on VG150~\cite{krishna2017visual} test set directly. 
    Our models trained on COCO captions are used as pre-trained models for OvR-SGG and OvD+R-SGG settings.
    }    
    \setlength\tabcolsep{5.3pt}
    \scalebox{0.94}{
    \begin{tabular}{|rl||cc|ccc|}
    \thickhline
    \rowcolor{mygray}
    \multicolumn{2}{|c||}{SGG model} &  Backbone & Grounding & R@20  & R@50  & R@100  \\
    \hline
    \hline
    LSWS~\cite{yelinguistic} &$_\text{CVPR'21}$& -& - & - & 3.28 & 3.69 \\ 
    MOTIFS~\cite{zellers2018neural}&$_\text{CVPR'18}$& - & Li \etal~\cite{li2022integrating} & 5.02 & 6.40 & 7.33 \\
    Uniter~\cite{chen2020uniter}&$_\text{ECCV'20}$& - & SGNLS \cite{zhong2021learning} & - & 5.80 & 6.70 \\
    Uniter~\cite{chen2020uniter}&$_\text{ECCV'20}$ & - & Li \etal  \cite{li2022integrating} & 5.42 & 6.74 & 7.62 \\
    \hline
    $\text{VS}^3$~\cite{zhang2023learning} &$_\text{CVPR'23}$&  \multirow{3}[0]{*}{Swin-T} & GLIP-L \cite{li2022grounded}  & 5.59 & 7.30 & 8.62 \\
    OvSGTR~\cite{chen2024expanding}& $_\text{ECCV'24}$ & & Grounding DINO \cite{liu2023grounding} & {6.61} &{8.92} &{10.90} \\ 
     \textbf{INOVA} (\textbf{Ours}) & & & Grounding DINO \cite{liu2023grounding} & \textbf{7.86} &\textbf{10.81} &\textbf{13.31} \\ 
     \hline
    OvSGTR~\cite{chen2024expanding}& $_\text{ECCV'24}$ & \multirow{2}[0]{*}{Swin-B} & Grounding DINO \cite{liu2023grounding} & {6.88} & {9.30} & {11.48} \\
     \textbf{INOVA} (\textbf{Ours})& & & Grounding DINO \cite{liu2023grounding} & \textbf{8.28} & \textbf{11.61} &\textbf{14.33} \\
    \thickhline
    \end{tabular}
    }
    \vspace{-0.5em}
    \label{tab:pretrain}
\end{table*}

\begin{figure}[!t]
    \centering
    \vspace{-0.3em}
    \includegraphics[width=1\linewidth]{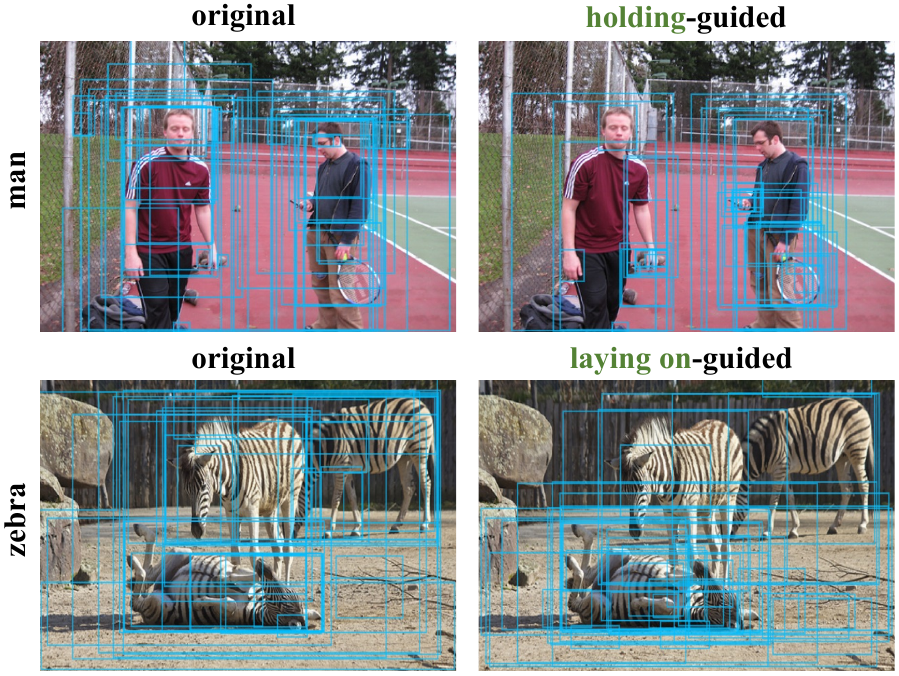}
    \vspace{-2.3em}
    \caption{Interaction-guided query selection.}
    \label{fig:iqs}
    \vspace{-1.2em}
\end{figure}

\textbf{Supervision Analysis.}
We investigated ITG's impact in the pre-training process (\cf Table~\ref{tab:pretrain}). As seen, models pre-trained on COCO~\cite{chen2015microsoft} captions with INOVA variants consistently outperform others, achieving \textbf{13.31}\% R@100 with Swin-T and \textbf{14.22}\% R@100 with Swin-B. These results demonstrate the effectiveness of incorporating ITG in the VLM pre-training process.

In addition, we visualized the object detection results from ITG and the original methods that solely use object categories for detection. As displayed in Figure~\ref{fig:framework}, the original method produces redundant objects, complicating the identification of subject-object interactions. For instance, given the ``$\langle$\texttt{people}, \texttt{ride}, \texttt{bike}$\rangle$'' triplet, the baseline detects multiple instances of ``\texttt{people}'' and ``\texttt{bike}'', obscuring the interaction. In contrast, ITG leverages bidirectional interaction prompts and attention mechanisms to accurately localize the interaction-relevant objects. A similar enhancement is observed in the ``$\langle$\texttt{bikes}, \texttt{on}, \texttt{boat}$\rangle$'' triplet, where ITG focuses on interaction-relevant entities. 

\textbf{Query Visualization.} To demonstrate the effectiveness of IQS, we visualized the top-50 selected queries in Figure~\ref{fig:iqs}. As seen, the original approach makes no distinction between instances within the same category, such as ``\texttt{man}'' or ``\texttt{zebra}'', resulting in both interacting and non-interacting instances receiving a similar number of queries. This indiscriminate query generation increases the likelihood of incorrect matches during bipartite graph matching, as irrelevant regions compete with interaction-relevant instances. Conversely, IQS prioritizes queries for interacting instances (``man holding'' or ``zebra laying on'' in Figure~\ref{fig:iqs}), increasing discrimination among objects with the same categories.

\section{Conclusion}

This work presents an interaction-aware framework INOVA for OVSGG. Unlike previous works that treat all objects equally, INOVA emphasizes the distinction between interacting and non-interacting objects, which is crucial for exact relation recognition. By adopting interaction-aware target generation, interaction-guided query selection, and interaction-consistent knowledge distillation, INOVA effectively mitigates issues like mismatched relation pairs and irrelevant object interference. INOVA shows significant improvements across two mainstream OVSGG benchmarks. We anticipate that INOVA will not only set a new standard for OVSGG but also inspire further exploration of interaction-driven methodologies in VLMs for more accurate scene understanding.

\section*{Impact Statement}
This paper presents work whose goal is to improve open-vocabulary scene graph generation. While our method, INOVA, focuses on technical advancements in introducing interaction-aware mechanisms, we acknowledge the broader implications of such technology. Enhanced scene graph generation could enable more robust applications in areas like assistive technologies, autonomous systems, and content-based image retrieval. However, as with many ML systems, biases in training data or deployment contexts could propagate unintended societal effects. We encourage future work to rigorously evaluate fairness and robustness when applying such models in critical domains.

\bibliography{ref}
\bibliographystyle{icml2025}



\end{document}